\documentclass[runningheads]{llncs}

\usepackage{eccv}

\usepackage{eccvabbrv}

\usepackage{graphicx}
\usepackage{booktabs}

\usepackage[accsupp]{axessibility}  

\usepackage{hyperref}

\usepackage{orcidlink}

\usepackage{graphicx}
\usepackage{amsmath}
\usepackage{amssymb}
\usepackage{booktabs}
\usepackage{multirow}
\usepackage{diagbox}
\usepackage{color}
\usepackage{booktabs}
\usepackage{multirow}
\usepackage{bbding}
\usepackage{makecell}
\usepackage{pifont}

\begin{document}

\title{SCPNet: Unsupervised Cross-modal \\ Homography Estimation via \\ Intra-modal Self-supervised Learning} 

\titlerunning{SCPNet}

\author{Runmin Zhang\inst{1}$^*$ \and
Jun Ma\inst{2,1}$^*$ \and
Si-Yuan Cao\inst{2,1}$^*$$^\dagger$  \and
Lun Luo\inst{3} \and \\
Beinan Yu\inst{1} \and
Shu-Jie Chen\inst{4} \and
Junwei Li\inst{1} \and
Hui-Liang Shen\inst{1}}

\authorrunning{R. Zhang et al.}

\institute{
College of Information Science and Electronic Engineering, Zhejiang University \and
Ningbo Innovation Center, Zhejiang University \and
HAOMO.AI Technology Co., Ltd. \and
Zhejiang GongShang University \\
\footnotetext{$^*$ Equal Contributions. $^\dagger$ Corresponding author. (cao\_siyuan@zju.edu.cn)}
}

\maketitle

\begin{abstract}
We propose a novel unsupervised cross-modal homography estimation framework based on intra-modal \textbf{S}elf-supervised learning, \textbf{C}orrelation, and consistent feature map \textbf{P}rojection, namely SCPNet. The concept of intra-modal self-supervised learning is first presented to facilitate the unsupervised cross-modal homography estimation. The correlation-based homography estimation network and the consistent feature map projection are combined to form the learnable architecture of SCPNet, boosting the unsupervised learning framework. SCPNet is the first to achieve effective unsupervised homography estimation on the satellite-map image pair cross-modal dataset, GoogleMap, under [-32,+32] offset on a $128 \times 128$ image, leading the supervised approach MHN by 14.0\% of mean average corner error (MACE). We further conduct extensive experiments on several cross-modal/spectral and manually-made inconsistent datasets, on which SCPNet achieves the state-of-the-art (SOTA) performance among unsupervised approaches, and owns 49.0\%, 25.2\%, 36.4\%, and 10.7\% lower MACEs than the supervised approach MHN. Source code is available at \url{https://github.com/RM-Zhang/SCPNet}.
\end{abstract}

\section{Introduction}
\label{sec:intro}

Homography estimation aims to compute the global perspective transform among images. Present supervised homography estimation approaches \cite{detone2016deep,le2020deep,zhao2021deep,shao2021localtrans,cao2022iterative,cao2023recurrent} can usually handle the homography estimation task under large offsets and modality gaps. However, in real applications, the homography deformation between images is usually unknown, especially for the cross-modal images captured by different devices or at various times \cite{yasuma2010generalized,chen2017normalized}. Therefore, unsupervised cross-modal homography estimation is vital for real-world tasks such as multi-spectral image fusion \cite{ying2021unaligned,zhou2019integrated}, multi-modal image restoration \cite{marivani2022designing,dharejo2022multimodal}, and GPS denied navigation \cite{goforth2019gps, zhao2021deep}.

\begin{figure}[t]
	\centering
	\includegraphics[scale=0.42]{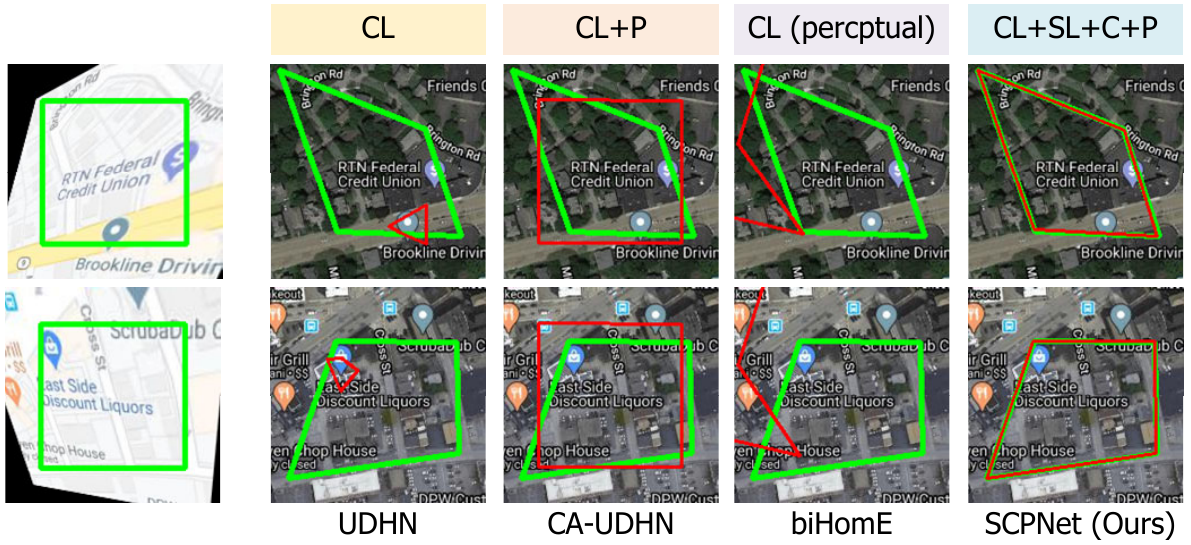}
	\caption{Unsupervised homography estimation results of UDHN \cite{nguyen2018unsupervised}, CA-UDHN \cite{zhang2020content}, biHomE \cite{koguciuk2021perceptual}, and our SCPNet on GoogleMap dataset under [-32,+32] offset. CL denotes the common cross-modal intensity-based learning, SL denotes the intra-modal self-supervised learning, C denotes correlation, and P denotes consistent feature map projection. CL (perceptual) means the cross-modal intensity-based learning is conducted by the perceptual loss. {\color{green}Green} polygons denote the ground-truth homography deformation from $\mathbf{I}_\mathrm{B}$ (map) to $\mathbf{I}_\mathrm{A}$ (satellite). {\color{red}Red} polygons denote the estimated homography deformation using different algorithms on $\mathbf{I}_\mathrm{A}$ (satellite). Different from the previous works that only adopt cross-modal intensity-based learning, SCPNet introduces intra-modal self-supervised learning as extra supervision and has a special architecture based on correlation and consistent feature map projection, leading to successful unsupervised cross-modal homography learning under large offsets and modality gaps. }
	\label{fig:first}
\end{figure}

For the above reasons, unsupervised deep homography estimation has raised growing interest. Nguyen \etal \cite{nguyen2018unsupervised} trained a deep homography estimation network in an unsupervised manner by comparing the pixel intensity of warped source image and target image. Wang \etal 
\cite{wang2019self} constrained the intensity loss in a cyclic manner that further improves the homography estimation accuracy. To cope with the illumination change, several works have been presented \cite{zhang2020content, ye2021motion} to achieve unsupervised homography estimation by introducing the feature representation for both homography estimation and consistency supervision. Based on the above two works, Hong \etal  \cite{hong2022unsupervised} adopted the GAN \cite{goodfellow2014generative} model to improve the supervision of feature similarity. However, most of the above approaches focus only on cross-modal intensity-based learning, and can only separately address either large offsets or modality gaps \cite{koguciuk2021perceptual}.

To cope with the above problem, in this paper, we propose a novel unsupervised cross-modal homography estimation framework, namely SCPNet, which adopts intra-modal \textbf{S}elf-supervised learning, \textbf{C}orrelation, and consistent feature map \textbf{P}rojection. As illustrated in Fig. \ref{fig:first}, different from the previous unsupervised works that only adopt cross-modal intensity-based learning, SCPNet introduces intra-modal self-supervised learning as extra supervision and has a special architecture based on correlation and consistent feature map projection. It is observed that SCPNet achieves successful unsupervised homography estimation on the cross-modal data under such large offsets, while the others cannot. 

The intra-modal self-supervised learning lays the foundation of our SCPNet, which mines the two-branch self-supervised information via applying simulated homography within the two modalities. The network with shared weights is trained simultaneously by the two-branch self-supervised learning. According to the ablation on GoogleMap, simply using the intra-modal self-supervised learning, our SCPNet can produce converged training, even without the cross-modal intensity-based learning in \cite{nguyen2018unsupervised,wang2019self,zhang2020content, ye2021motion}. On the contrary, only using the cross-modal intensity-based learning fails to converge on such a large modality gap and offset. The two learning strategies are combined to form the final supervision of SCPNet. The correlation and consistent feature map projection have been separately employed in many previous homography estimation frameworks \cite{shao2021localtrans,cao2022iterative,zhang2020content,ye2021motion,hong2022unsupervised}. However, the strategy and effectiveness of combining them to form an effective unsupervised cross-modal homography estimation framework haven't ever been investigated. The above two parts form the powerful learnable architecture of our SCPNet, which also boosts the unsupervised training framework.

To the best of our knowledge, SCPNet is the first method that achieves effective unsupervised homography estimation on such a large offset (offset range of [-32,+32] on a $128\times128$ image) and modality gap (GoogleMap \cite{zhao2021deep} of satellite-map image pairs as in Fig. \ref{fig:first}), outperforming the supervised approach MHN \cite{le2020deep} by 14.0\% of mean average corner error (MACE). We further evaluate our SCPNet on Flash/no-flash \cite{he2014saliency} cross-modal dataset, Harvard \cite{chakrabarti2011statistics} and RGB/NIR \cite{brown2011multi} cross-spectral datasets, and PDS-COCO \cite{koguciuk2021perceptual} manually-made inconsistent dataset, which also achieves the state-of-the-art (SOTA) performance among unsupervised approaches. In summary, our contributions are as follows: 
\begin{itemize}
	\item{We propose SCPNet, a novel unsupervised cross-modal homography estimation framework, which combines three key components, including intra-modal self-supervised learning, correlation, and consistent feature map projection. SCPNet ranks top in the unsupervised homography estimation on cross-modal/spectral and manually-made inconsistent data under large offsets. }
	\item{The concept of intra-modal self-supervised learning is devised to support the unsupervised learning framework, which mines the two-branch self-supervised information via applying simulated homography within the two modalities. By simultaneously training the weight-shared network using the two-branch self-supervised learning, the homography estimation knowledge can be generalized from intra-modal to cross-modal.}
	\item{We combine the correlation and consistent feature map projection to form a powerful unsupervised learning network architecture of SCPNet. The correlation constrains the network to learn a clearer knowledge that can be generalized from intra-modal to cross-modal. The projected consistent feature map can monitor both the cross-modal homography estimation and cross-modal consistent latent space projection, which will further improve the estimation accuracy.}
\end{itemize}

\section{Related Work}

\textbf{Traditional Approaches.} The most widely used traditional homography estimation approaches, namely feature-based approaches, typically involve three key steps: feature extraction, feature matching, and homography estimation \cite{shao2021localtrans}. Commonly used feature extraction approaches include SIFT \cite{lowe2004distinctive}, SURF \cite{bay2006surf}, and ORB \cite{mur2015orb}. Popular homography estimation techniques include DLT \cite{dubrofsky2009homography}, RANSAC \cite{fischler1981random}, IRLS \cite{holland1977robust}, and MAGSAC \cite{barath2019magsac}. To further improve the robustness of cross-modal feature extraction, some approaches such as LGHD \cite{aguilera2015lghd}, RIFT \cite{li2019rift}, and DASC \cite{kim2015dasc} have been presented. The above approaches achieve reliable homography estimations under moderate intensity variance and deformation but may produce unsatisfactory results dealing with cross-modal images under large offsets \cite{cao2020boosting,cao2022iterative, le2020deep, shao2021localtrans}.

\textbf{Supervised Approaches.} DeTone \etal \cite{detone2016deep} first introduced the end-to-end homography estimation network DHN. To further improve the accuracy of homography estimation, many approaches have been subsequently presented. For example, MHN \cite{le2020deep} used a multi-scale network concatenation and DLKFM \cite{zhao2021deep} adopted deep Lucas-Kanade iteration. Furthermore, LocalTrans \cite{shao2021localtrans} trained a multi-scale local transformer, IHN \cite{cao2022iterative} employed deep learnable iteration, and RHWF \cite{cao2023recurrent} combined homography-guided image warping and focus transformer, \etc. However, obtaining the ground-truth is often difficult and costly, making it challenging for supervised learning approaches to be widely applicable in practice.

\textbf{Unsupervised Approaches.} Nguyen \etal \cite{nguyen2018unsupervised} trained the homography estimation network using pixel-level photometric loss in an unsupervised manner. Based on this pioneering work, Wang \etal \cite{wang2019self} added extra supervision by the invertibility constraints. Zhang \etal \cite{zhang2020content} presented CA-UDHN to depict the similarity in feature space instead of pixel space. However, CA-UDHN has poor robustness for images with large viewpoint changes \cite{koguciuk2021perceptual}. Koguchiuk \etal \cite{koguciuk2021perceptual} then expanded CA-UDHN with perceptual loss \cite{johnson2016perceptual}, which improves the robustness of unsupervised training of deep homography estimation under large intensity and viewpoint changes. Furthermore, Ye \etal \cite{ye2021motion} introduced feature identity loss to enforce the image feature to be warp-equivalent, and proposed a homography flow representation. Besides the above approaches, some unsupervised techniques such as NeMAR \cite{arar2020unsupervised}, UMF-CMGR \cite{Wang_2022_IJCAI}, and RFNet \cite{xu2022rfnet} use modality transfer networks to migrate one modality to another, achieving unsupervised cross-modal/spectral motion estimation.

\section{Pilot Experiments and Finding}
\label{subsec:pilot}

We first denote the image pair from modality A and B as $\mathbf{I}_\mathrm{A}$ and $\mathbf{I}_\mathrm{B}$, with homography deformation between them. To train a homography estimation network in a supervised manner, the objective of network training can be formulated as 
\begin{equation}
	\mathop{\arg\min}_{\theta}\mathcal{L}_\mathrm{S}\big(\phi_\theta(\mathbf{I}_\mathrm{A},\mathbf{I}_\mathrm{B}), \mathbf{H}_\mathrm{GT}\big),
	\label{eq:supervised}
\end{equation}
where $\mathbf{H}_\mathrm{GT}$ denotes the ground-truth homography between the two images, $\phi_\theta$ the network, and $\theta$ the network parameters to be optimized. $\mathcal{L}_\mathrm{S}$ denotes the supervised loss, which is usually $L_2$ \cite{le2020deep} or $L_1$ \cite{shao2021localtrans,cao2022iterative} norm. However, in practical applications, ground-truth homography is generally difficult to obtain, especially for the cross-modal images captured by different devices or at various times \cite{yasuma2010generalized,chen2017normalized}. To cope with this difficulty, unsupervised homography estimation is then investigated, and the training for most of them \cite{nguyen2018unsupervised,zhang2020content,koguciuk2021perceptual,hong2022unsupervised} can be modeled as 
\begin{equation}
	\mathop{\arg\min}_{\theta}\mathcal{L}_\mathrm{C}\big(\mathbf{I}_\mathrm{A}, \mathcal{W}(\mathbf{I}_\mathrm{B}, \phi_\theta(\mathbf{I}_\mathrm{A},\mathbf{I}_\mathrm{B}))\big),
	\label{eq:cross}
\end{equation}
where $\mathcal{W}$ denotes the warping operation using the predicted homography $\phi_\theta(\mathbf{I}_\mathrm{A},\mathbf{I}_\mathrm{B})$, and $\mathcal{L}_\mathrm{C}$ denotes the cross-modal loss that monitoring the content similarity of the warped $\mathbf{I}_\mathrm{B}$ and $\mathbf{I}_\mathrm{A}$. The cross-modal intensity-based loss varies from the $L_1$ pixel-wise photometric loss \cite{nguyen2018unsupervised}, the $L_1$ similarity loss of the feature maps  \cite{zhang2020content,ye2021motion}, and the perceptual loss \cite{koguciuk2021perceptual}. Nevertheless, under large homography deformation and intensity variance, the above losses may fail, according to \cite{koguciuk2021perceptual} and our experiments. As the cross-modal image intensity similarity is generally highly non-convex \cite{cao2020boosting}, making the solution space of the loss function hard to optimize, the training process is prone to fall into non-convergent as demonstrated in \cite{koguciuk2021perceptual}.

\begin{figure}[t]
	\centering
	\includegraphics[scale=0.35]{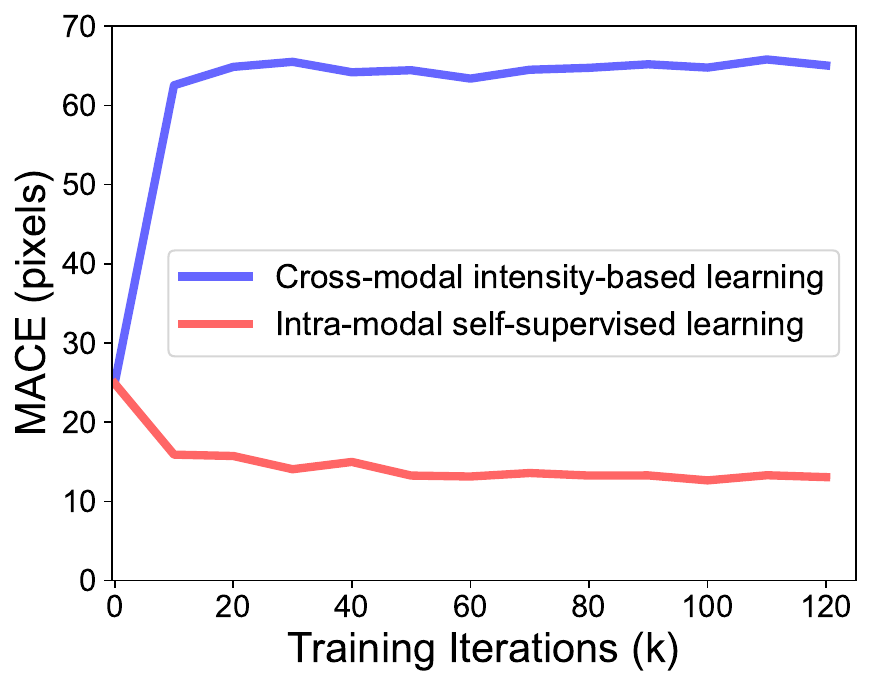}
	\caption{The cross-modal test MACEs of the network trained using intra-modal self-supervised learning and cross-modal intensity-based learning during the training iterations, respectively.}
	\label{fig:observation}
\end{figure}

Inspired by multitask learning \cite{caruana1997multitask}, which simultaneously tackles multiple tasks using a shared representation, we propose intra-modal self-supervised learning to achieve better supervision. Multitask learning implicitly learns task relationships within a shared representation through gradient aggregation \cite{caruana1997multitask}, and recent studies have shown that various tasks benefit from it \cite{doersch2017multi, hu2021unit, girdhar2022omnivore}. The motivation of our intra-modal self-supervised learning is to enhance the unsupervised learning process by introducing highly related extra tasks that provide direct supervision. While obtaining cross-modal ground-truth homography is challenging, intra-modal ground-truth homography can be easily generated by directly applying simulated deformations \cite{detone2016deep}. This allows the knowledge of homography transformation to be directly learned, rather than indirectly as in common cross-modal intensity-based learning. Additionally, the relationship between images from two modalities, such as mutual structures, is likely to be learned within the shared representation during two-branch intra-modal self-supervised learning. To validate the aforementioned statement, we train a weight-shared network to separately predict the homography within the two modalities under direct supervision from simulation, which can be expressed as
\begin{equation}
	\mathop{\arg\min}_{\theta} \mathcal{L}_\mathrm{S}\big(\phi_\theta(\mathbf{I}_\mathrm{A},\mathbf{I}'_\mathrm{A}), \mathbf{H}_\mathrm{GT,A}\big) + \mathcal{L}_\mathrm{S}\big(\phi_\theta(\mathbf{I}_\mathrm{B},\mathbf{I}'_\mathrm{B}), \mathbf{H}_\mathrm{GT,B}\big),
	\label{eq:selfsupervised}
\end{equation}
where $\mathbf{I}'_\mathrm{A}$ denotes the homography warped $\mathbf{I}_\mathrm{A}$ with the simulated ground-truth homography $\mathbf{H}_\mathrm{GT,A}$, and modality B in the same manner. We conduct this pilot experiment on the cross-modal dataset, GoogleMap \cite{zhao2021deep}, which is of large intensity and content difference, under [-32,+32] offset on a $128 \times 128$ image. The cross-modal test MACEs of the network trained using intra-modal self-supervised learning and common cross-modal intensity-based learning during the training iterations are illustrated in Fig. \ref{fig:observation}. Interestingly, we find that the network trained by intra-modal self-supervised learning has an evidently better cross-modal performance than the common cross-modal intensity-based learning trained one. Therefore, we can obtain the finding: \textbf{The cross-modal homography estimation can be indirectly facilitated by training the weight-shared network using the simulated transform within the two modalities.}

\section{SCPNet}

\begin{figure*}[t]
	\centering
	\includegraphics[scale=0.15]{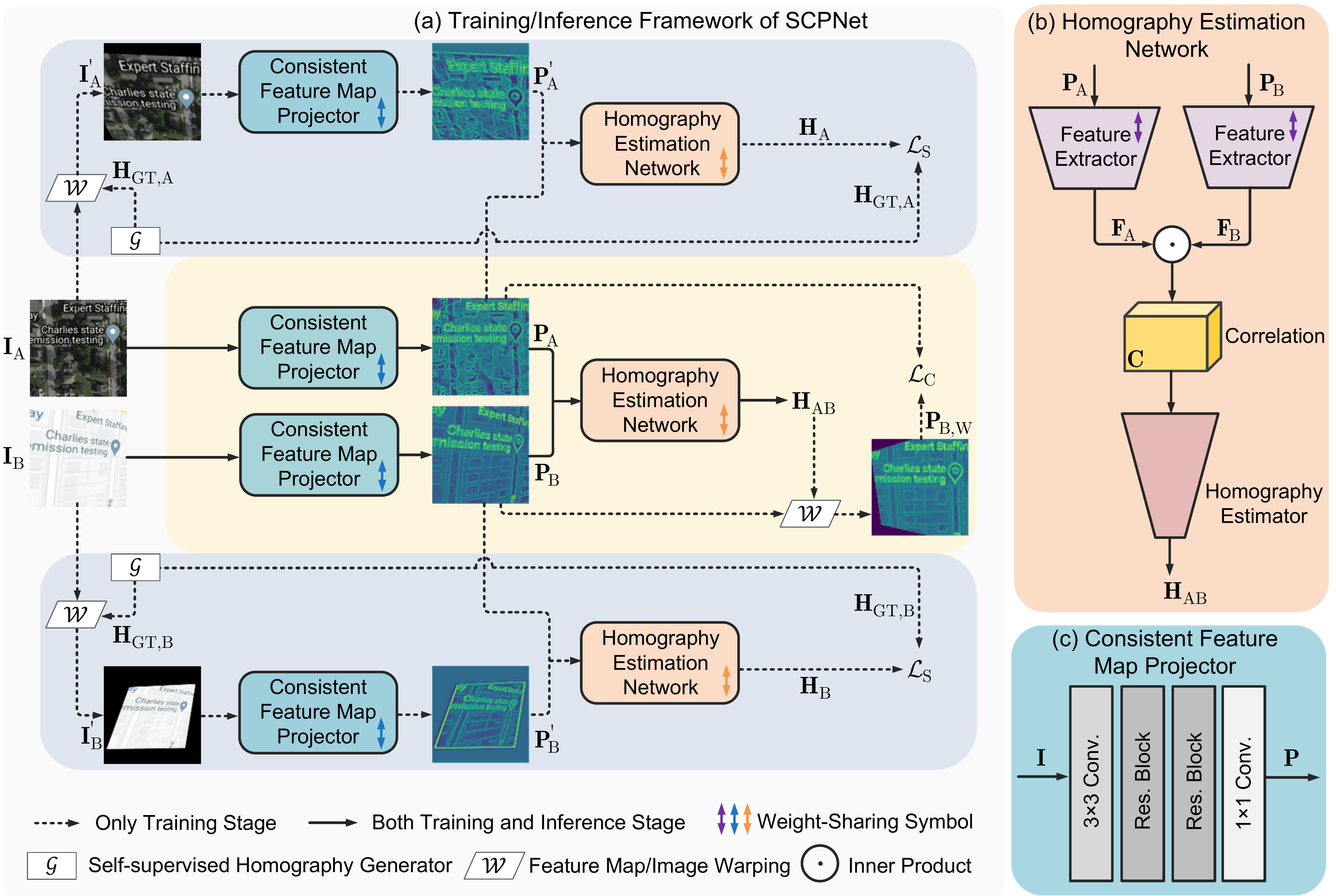}
	\caption{Schematic diagram of unsupervised cross-modal homography estimation framework using intra-modal \textbf{S}elf-supervised learning, \textbf{C}orrelation, and consistent feature map \textbf{P}rojection, namely SCPNet. (a) Overall structure and training/inference strategy of SCPNet. (b) Detailed illustration of the correlation-based homography estimation network. (c) Detailed structure of the consistent feature map projector.}
	\label{fig:method}
\end{figure*}

Based on the finding in Section \ref{subsec:pilot}, we hope to further design a network architecture and complement it with an appropriate training strategy. For this purpose, we propose the unsupervised cross-modal homography estimation framework that adopts intra-modal \textbf{S}elf-supervised learning, \textbf{C}orrelation, and consistent feature map \textbf{P}rojection, namely SCPNet. Fig. \ref{fig:method}(a) demonstrates the schematic diagram of the training and inference framework of SCPNet. Fig. \ref{fig:method}(b) and \ref{fig:method}(c) show the two learnable modules that form the architecture of SCPNet. Considering that the learnable modules and the training strategy are coupled and mutually promote each other, in the following section, we will follow the idea of building a powerful unsupervised learning framework based on the finding of intra-modal self-supervised learning to demonstrate our SCPNet.

\subsection{Correlation-based Homography Estimation Network}
\label{subsec:corr}

Similar to the previous unsupervised network architectures \cite{nguyen2018unsupervised,wang2019self,zhang2020content,ye2021motion,koguciuk2021perceptual}, the finding in Section \ref{subsec:pilot} is obtained by concatenating the image pairs in the channel dimension and expecting the network to directly predict the homography. The knowledge of homography estimation is implicitly learned without any constraint or hint, and hence the potential of our intra-modal self-supervised learning might not be fully explored.

With the above consideration, we alter to construct the homography estimation network with correlation. The architecture of the correlation-based homography estimation network is illustrated in Fig. \ref{fig:method}(b). The feature extractor with shared weights produces the features of the two modalities, namely $\mathbf{F}_\mathrm{A}$ and $\mathbf{F}_\mathrm{B}$. The correlation is realized by computing the inner product of $\mathbf{F}_\mathrm{A}$ and $\mathbf{F}_\mathrm{B}$ around a local area, which can be expressed as
\begin{equation}
	\begin{aligned}
		\mathbf{C}(\mathbf{x},\mathbf{r}) = \mathrm{ReLU}(\mathbf{F}_\mathrm{A}(\mathbf{x})^{\mathsf{T}} \mathbf{F}_\mathrm{B}(\mathbf{x} + \mathbf{r})),\ \ \ \ \|\mathbf{r}\|_{\infty} \leq R
		\label{eq:correlation_volume}
	\end{aligned}
\end{equation}
where $R$ controls the radius of each local area. The correlation is then sent into the homography estimator to conduct the homography prediction. The structural details of the homography estimation network can be found in the supplementary material. By adopting correlation, the homography estimation network is clarified into the weight-shared feature extractor, correlation computation, and the homography estimator. Under the intra-modal self-supervised learning, each of the above parts is constrained to learn a clearer knowledge that can be generalized to cross-modal: \textbf{1)} the feature extractor is constrained to produce feature representations that are effective for correlation, which is indirectly enforced to share the intra-modal self-supervised knowledge to cross-modal; \textbf{2)} the similarity of features is explicitly encoded by correlation, which is unified among each modality; \textbf{3)} the knowledge of homography decoding is strictly defined by the correlation input, which is also unified. The intra-modal self-supervised learning of the network can then be formulated by
\begin{equation}
	\mathop{\arg\min}_{\xi} \mathcal{L}_\mathrm{S}\big(\psi_\xi(\mathbf{I}_\mathrm{A},\mathbf{I}'_\mathrm{A}), \mathbf{H}_\mathrm{GT,A}\big) + \mathcal{L}_\mathrm{S}\big(\psi_\xi(\mathbf{I}_\mathrm{B},\mathbf{I}'_\mathrm{B}), \mathbf{H}_\mathrm{GT,B}\big),
	\label{eq:selfsupervised_corr}
\end{equation}
where $\psi_\xi$ denotes the correlation-based homography estimation network, with the parameters to be optimized by $\xi$.

\subsection{Consistent Feature Map Projector}
\label{subsec:consis}

After introducing intra-modal self-supervised learning, we then consider bringing in valid cross-modal supervision to further improve the estimation accuracy. As discussed in Section \ref{subsec:pilot}, directly applying intensity-based supervision in Eq. \ref{eq:cross} on cross-modal images with severe content differences is infeasible. To cope with the problem, we introduce consistent feature map projection to assist the intensity-based cross-modal supervision, which projects the input images from an intensity-variant space to an intensity-invariant latent space. The architecture of the consistent feature map projector is illustrated in Fig. \ref{fig:method}(c). The input image is processed with a convolutional block of kernel size $3 \times 3$ first. The produced feature map is then processed by two residual blocks. Finally, the feature map with a higher number of channels is projected into the consistent feature map of $1$ channel by a $1 \times 1$ convolutional block.

Boosted by the consistent feature map projector, the cross-modal intensity-based training can be expressed as
\begin{equation}
	\mathop{\arg\min}_{\xi, \zeta}\mathcal{L}_\mathrm{C}\big(\delta_\zeta(\mathbf{I}_\mathrm{A}), \mathcal{W}(\delta_\zeta(\mathbf{I}_\mathrm{B}), \psi_\xi(\delta_\zeta(\mathbf{I}_\mathrm{A}),\delta_\zeta(\mathbf{I}_\mathrm{B})))\big),
	\label{eq:cross_proj}
\end{equation}
where $\delta_\zeta$ denotes the consistent feature map projector, with $\zeta$ denoting its learnable parameters. We note that, with the consistent feature map projector, the cross-modal intensity-based learning not only supervises the cross-modal homography estimation but also makes the projected feature maps as similar as possible, which will further boost the estimation accuracy.

\subsection{Training/Inference Framework}
\label{sub:final}
Now that we have separately introduced the intra-modal self-supervised learning, the correlation-based homography estimation network, and the consistent feature map projector with cross-modal intensity-based learning. The complete framework of SCPNet can be determined by combining the above learning strategies and modules. The training framework of SCPNet contains two self-supervised learning branches and one cross-modal learning branch, which is the most significant difference compared to the previous approaches. The three branches apply simultaneously supervision on the weight-shared learnable modules as demonstrated in Fig. \ref{fig:method}(a). For better illustration, the projected consistent feature maps are denoted by $\mathbf{P}_\mathrm{A}=\delta_\zeta(\mathbf{I}_\mathrm{A})$, $\mathbf{P}_\mathrm{B}=\delta_\zeta(\mathbf{I}_\mathrm{B})$, and the warped $\mathbf{P}_\mathrm{B}$ by 
$\mathbf{P}_\mathrm{B,W}=\mathcal{W}(\delta_\zeta(\mathbf{I}_\mathrm{B}), \psi_\xi(\delta_\zeta(\mathbf{I}_\mathrm{A}),\delta_\zeta(\mathbf{I}_\mathrm{B})))$. The predicted cross-modal homography is denoted by $\mathbf{H_\mathrm{AB}=\psi_\xi(\delta_\zeta(\mathbf{I}_\mathrm{A}),\delta_\zeta(\mathbf{I}_\mathrm{B}))}$, and intra-modal ones by $\mathbf{H}_\mathrm{A}=\psi_\xi(\delta_\zeta(\mathbf{I}_\mathrm{A}),\delta_\zeta(\mathbf{I}'_\mathrm{A}))$ and $\mathbf{H}_\mathrm{B}=\psi_\xi(\delta_\zeta(\mathbf{I}_\mathrm{B}),\delta_\zeta(\mathbf{I}'_\mathrm{B}))$. As mentioned in Section \ref{subsec:pilot}, for the two self-supervised branches, the input $\mathbf{I}_\mathrm{A}$ and $\mathbf{I}_\mathrm{B}$ are separately deformed and trained under the direct supervision of the simulated homography $\mathbf{H}_\mathrm{GT,A}$ and $\mathbf{H}_\mathrm{GT,B}$. Meanwhile, cross-modal intensity-based learning is conducted by applying supervision on the projected consistent feature map $\mathbf{P}_\mathrm{A}$ and warped one $\mathbf{P}_\mathrm{B,W}$. The correlation-based homography estimation network and consistent feature map projector are both absorbed to form the network. Finally, the entire unsupervised cross-modal learning framework can be formulated as
\begin{equation}
	\begin{aligned}
		\mathop{\arg\min}_{\xi,\zeta} \ \   & \mathcal{L}_\mathrm{C}  \big(\delta_\zeta(\mathbf{I}_\mathrm{A}), \mathcal{W} ( \delta_\zeta(\mathbf{I}_\mathrm{B}), \psi_\xi ( \delta_\zeta(\mathbf{I}_\mathrm{A}), \delta_\zeta(\mathbf{I}_\mathrm{B})))\big)\\ 
		+&\lambda\mathop{\mathcal{L}_\mathrm{S}}_{}\big(\psi_\xi(\delta_\zeta(\mathbf{I}_\mathrm{A}),\delta_\zeta(\mathbf{I}'_\mathrm{A})), \mathbf{H}_\mathrm{GT,A}\big)_{\ }\\ 
		+ &\lambda\mathop{\mathcal{L}_\mathrm{S}}_{}\big(\psi_\xi(\delta_\zeta(\mathbf{I}_\mathrm{B}),\delta_\zeta(\mathbf{I}'_\mathrm{B})), \mathbf{H}_\mathrm{GT,B}\big).
		\label{eq:full}
	\end{aligned}
\end{equation}
We note that once combined, the correlation can also facilitate the projected consistent feature map to have clear contents with the promotion of the cross-modal intensity learning, which will be discussed in Section \ref{subsec:ablation}. As for the inference phase, only the cross-modal prediction branch of SCPNet functions.

\subsection{Loss Function and Implementation Details} 
For the intra-modal self-supervised loss, we parameterize the homography matrix by the offsets of four corner points to stabilize the training \cite{detone2016deep, cao2022iterative, cao2023recurrent}. We use the $L_1$ norm on the differences between the predicted offsets $\mathbf{O} \in \mathbb{R}^{2\times2\times2}$ and the ground-truth $\mathbf{O}_\mathrm{GT} \in \mathbb{R}^{2\times2\times2}$, which can be formulated as:
\begin{equation}
	\mathcal{L}_\mathrm{S} = \|\mathbf{O} - \mathbf{O}_\mathrm{GT}\|_1.
\end{equation}
The homography parameterization using offsets of four corner points can be found in the supplementary material.

We set the cross-modal intensity-based loss as follows:
\begin{equation}
	\mathcal{L}_\mathrm{C} = \frac{\|\mathbf{P}_\mathrm{A} -  \mathbf{P}_\mathrm{B,W}\|_1}{\|\mathbf{P}_\mathrm{A} -  \mathbf{P}_\mathrm{B}\|_1},
	\label{eq:lc}
\end{equation}
where the numerator minimize the differences between the consistent feature map $\mathbf{P}_\mathrm{A}$ and the warped one $\mathbf{P}_\mathrm{B,W}$, while the denominator maximizing the differences between $\mathbf{P}_\mathrm{A}$ and $\mathbf{P}_\mathrm{B}$, which can prevent invalid feature map projection.

We set $\lambda=0.1$ in Eq. \ref{eq:full} during training. We use the AdamW \cite{loshchilov2017decoupled} optimizer, with the maximum learning rate of $4 \times 10^{-4}$ for the network training. The batch size is set to $8$, with a total of $120000$ training iterations.

\section{Experiments}
\label{sec:experiments}

\subsection{Datasets and Experimental Settings}
\textbf{Datasets.} We evaluate our SCPNet on cross-modal datasets including GoogleMap \cite{zhao2021deep} and Flash/no-flash \cite{he2014saliency}, cross-spectral datasets including Harvard \cite{chakrabarti2011statistics} and RGB/NIR \cite{brown2011multi}, together with the manually-made inconsistent dataset PDS-COCO \cite{koguciuk2021perceptual}. The GoogleMap dataset contains satellite images and the corresponding map images, which can be used for navigation and geolocation. We use the same training and test data splitting as in \cite{zhao2021deep}. The Flash/no-flash dataset contains 120 pairs of images that are with and without flash. We randomly select 60 image pairs for training and 60 for testing. For multi-spectral data, the Harvard dataset contains 77 real-world image scenes, with each scene containing 31 band images. We take the 16th band image of each scene as the reference image and form a cross-spectral image pair with the image of each remaining band respectively. The training and test data are divided by different scenes of 1170 and 1140 image pairs. For the RGB/NIR dataset, we use 103 pairs of images for training and 153 pairs for testing. PDS-COCO artificially simulates random combined changes in brightness, contrast, saturation, and hue noise to the MS-COCO dataset \cite{lin2014microsoft}. We use the same training and test splitting as the MS-COCO dataset.

\textbf{Experimental Settings.} The homography deformation is generated in the same way as \cite{cao2023recurrent,cao2022iterative, detone2016deep, zhao2021deep}, which randomly perturbs the four corner points of a $128\times128$ image. Unless otherwise stated, the perturbation range is set to [-32, +32]. We adopt the mean average corner error (MACE) for homography accuracy evaluation. Lower MACE indicates higher accuracy.

\textbf{Comparison Approaches.} We evaluate SCPNet on cross-modal and cross-spectral datasets with handcrafted approaches including SIFT \cite{lowe2004distinctive}, ORB \cite{rublee2011orb}, DASC \cite{kim2015dasc}, RIFT \cite{li2019rift}, unsupervised approaches including UDHN \cite{nguyen2018unsupervised}, CA-UDHN \cite{zhang2020content}, biHomE \cite{koguciuk2021perceptual}, BasesHomo \cite{ye2021motion}, UMF-CMGR \cite{Wang_2022_IJCAI}, and supervised approaches including DHN \cite{detone2016deep}, MHN \cite{le2020deep}, LocalTrans \cite{shao2021localtrans}, IHN \cite{cao2022iterative}, RHWF \cite{cao2023recurrent}. For SIFT, ORB, DASC, and UMF-CMGR, we choose RANSAC \cite{fischler1981random} as their homography estimation and outlier rejection algorithm. In addition, UMF-CMGR is an image fusion approach based on registration, and we use the registration network part for comparison. We also tried to compare with the unsupervised approaches MU-Net \cite{ye2022multiscale} and NeMAR \cite{arar2020unsupervised}, but according to our experiments, neither of them performs successful homography estimation. To make a more comprehensive comparison, we also evaluate our SCPNet on PDS-COCO \cite{koguciuk2021perceptual}.

\subsection{Ablation}
\label{subsec:ablation}

\textbf{Ablation Study on GoogleMap Dataset.} We conduct extensive ablation studies on the architecture and supervision of our SCPNet by evaluating the mean average corner error (MACE), as shown in Table \ref{tab:ablation}. Using only cross-modal intensity-based learning for supervision leads to non-convergence or unsatisfactory results (Settings 1–4). In contrast, our intra-modal self-supervised learning achieves superior performance (Settings 5–9). Moreover, the results of SCPNet show gradual improvement as additional ablation components are incorporated.

\begin{table}[tb]
	\caption{Ablation study of SCPNet. NC denotes the training is not converged. Self denotes intra-modal self-supervised learning, projection denotes consistent feature map projection, and cross denotes cross-modal intensity-based learning.}
	\renewcommand\arraystretch{1.2}
	\renewcommand\tabcolsep{8pt}{}
	\centering
	{
		\begin{tabular}{c|cccc|c}
			\hline
			\specialrule{0.1em}{0em}{0em}
			Setting & Self & Correlation & Projection & Cross & MACE$\downarrow$\\
			\hline
			1 & \scalebox{0.75}{\XSolidBrush} & \scalebox{0.75}{\XSolidBrush} & \scalebox{0.75}{\XSolidBrush} & \scalebox{0.75}{\Checkmark} & NC \\
			2 & \scalebox{0.75}{\XSolidBrush} & \scalebox{0.75}{\Checkmark} & \scalebox{0.75}{\XSolidBrush} & \scalebox{0.75}{\Checkmark} & NC \\
			3 & \scalebox{0.75}{\XSolidBrush} & \scalebox{0.75}{\XSolidBrush} & \scalebox{0.75}{\Checkmark} & \scalebox{0.75}{\Checkmark} & 24.64 \\
			4 & \scalebox{0.75}{\XSolidBrush} & \scalebox{0.75}{\Checkmark} & \scalebox{0.75}{\Checkmark} & \scalebox{0.75}{\Checkmark} & 24.80 \\
			\hline
			5 & \scalebox{0.75}{\Checkmark} & \scalebox{0.75}{\XSolidBrush} & \scalebox{0.75}{\XSolidBrush} & \scalebox{0.75}{\XSolidBrush} & 13.06 \\
			6 & \scalebox{0.75}{\Checkmark} & \scalebox{0.75}{\Checkmark} & \scalebox{0.75}{\XSolidBrush} & \scalebox{0.75}{\XSolidBrush} & 9.68 \\
			7 & \scalebox{0.75}{\Checkmark} & \scalebox{0.75}{\XSolidBrush} & \scalebox{0.75}{\Checkmark} & \scalebox{0.75}{\XSolidBrush} & 10.01 \\
			8 & \scalebox{0.75}{\Checkmark} & \scalebox{0.75}{\Checkmark} & \scalebox{0.75}{\Checkmark} & \scalebox{0.75}{\XSolidBrush} & 7.70 \\
			9 & \scalebox{0.75}{\Checkmark} & \scalebox{0.75}{\Checkmark} & \scalebox{0.75}{\Checkmark} & \scalebox{0.75}{\Checkmark} & 4.35 \\
			\hline
			\specialrule{0.1em}{0em}{0em}
		\end{tabular}
	}
	\label{tab:ablation}
\end{table}

\begin{figure}[tb]
	\centering
	\includegraphics[scale=0.095]{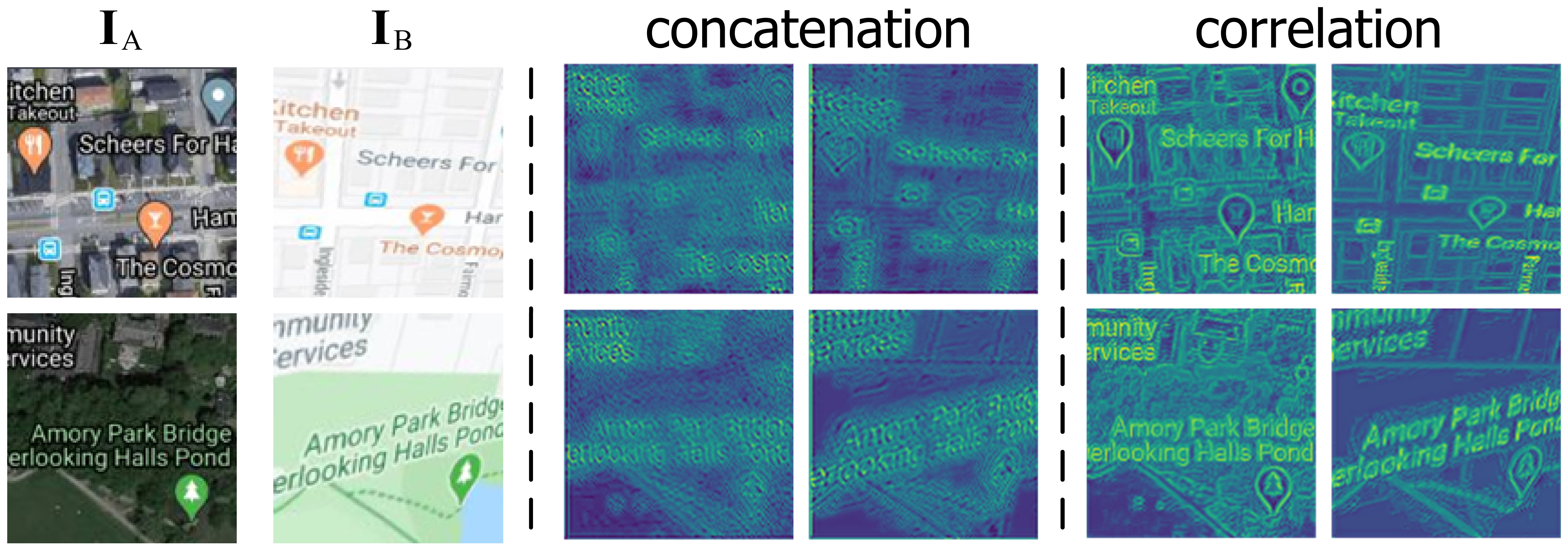}
	\caption{Comparison of the consistent feature maps produced by concatenation and correlation.}
	\label{fig:ablation_corr}
\end{figure}

\textbf{The Effectiveness of Correlation on Consistent Feature Map Projection.} We further show the consistent feature maps produced by concatenation and correlation architecture in Fig. \ref{fig:ablation_corr}. It is observed that the correlation visibly facilitates the consistent feature map generation by the direct constraint of the input feature maps using the inner product. Therefore, cross-modal intensity-based learning can then be boosted by high-quality feature maps. 

\begin{table}
	\caption{Quantitative results of our SCPNet and other approaches on cross-modal datasets. NC denotes the training is not converged. \textbf{Bold} indicates the best result among unsupervised methods.}
	\renewcommand{\tabcolsep}{4pt}{}
	\renewcommand\arraystretch{1.1}
	\centering
	\resizebox{\linewidth}{!}
	{
		\begin{tabular}{cc|cccc|cccc}
			\hline
			\specialrule{0.1em}{0em}{0em}
			\multicolumn{2}{c|}{Dataset} & \multicolumn{4}{c|}{GoogleMap} & \multicolumn{4}{c}{Flash/no-flash} \\
			\cline{1-2}
			\multicolumn{2}{c|}{Offset} & Easy & Moderate & Hard & Mean & Easy & Moderate & Hard & Mean \\
			\hline
			\multirow{4}{*}{Handcrafted} & SIFT \cite{lowe2004distinctive} & 19.17 & 23.87 & 29.04 & 24.53 & 14.61 & 18.69 & 23.85 & 19.53 \\
			\multirow{4}{*}{} & ORB \cite{rublee2011orb} & 19.11 & 23.9 & 29.02 & 24.52 & 16.91 & 22.44 & 27.01 & 22.63\\
			\multirow{4}{*}{} & DASC \cite{kim2015dasc} & 14.29 & 20.73 & 28.12 & 21.76 & 11.64 & 19.50 & 28.11 & 20.59 \\
			\multirow{4}{*}{} & RIFT \cite{li2019rift} & 10.43 & 15.46 & 21.93 & 16.55 & 11.22 & 13.95 & 21.66 & 16.21 \\
			\hline
			\multirow{6}{*}{Unsupervised} & UDHN \cite{nguyen2018unsupervised} & 18.63 & 21.55 & 26.89 & 22.84 & 16.27 & 21.27 & 24.85 & 21.20\\
			\multirow{6}{*}{} & CA-UDHN \cite{zhang2020content} & 19.31 & 23.92 & 29.10 & 24.61 & 16.01 & 21.54 & 25.14 & 21.32\\
			\multirow{6}{*}{} & biHomE \cite{koguciuk2021perceptual} & NC & NC & NC & NC & 8.24 & 12.56 & 14.04 & 11.86 \\
			\multirow{6}{*}{} & BasesHomo \cite{ye2021motion} & 19.43 & 23.97 & 28.66 & 24.49 & 19.45 & 24.73 & 29.66 & 25.12 \\
			\multirow{6}{*}{} & UMF-CMGR \cite{Wang_2022_IJCAI} & 19.22 & 24.01 & 29.02 & 24.60 & 17.99 & 22.43 & 28.40 & 23.49 \\
			\multirow{6}{*}{} & SCPNet (Ours) & \textbf{3.60} & \textbf{4.44} & \textbf{4.85} & \textbf{4.35} & \textbf{1.80} & \textbf{2.33} & \textbf{3.59} & \textbf{2.67} \\
			\hline
			\multirow{5}{*}{Supervised} & DHN \cite{detone2016deep} & 7.06 & 6.82 & 7.00 & 6.93 & 5.28 & 6.13 & 7.51 & 6.42 \\
			\multirow{5}{*}{} & MHN \cite{le2020deep} & 4.75 & 5.00 & 5.34 & 5.06 & 3.18 & 6.55 & 5.81 & 5.24 \\
			\multirow{5}{*}{} & LocalTrans \cite{shao2021localtrans} & 0.91 & 1.43 & 6.30 & 3.22 & 0.49 & 0.67 & 4.05 & 1.96 \\
			\multirow{5}{*}{} & IHN \cite{cao2022iterative} & 0.70 & 0.96 & 1.06 & 0.92 & 0.76 & 0.65 & 0.94 & 0.80 \\
			\multirow{5}{*}{} & RHWF \cite{cao2023recurrent} & 0.62 & 0.68 & 0.93 & 0.76 & 0.79 & 0.68 & 0.53 & 0.65 \\
			\hline
			\specialrule{0.1em}{0em}{0em}
		\end{tabular}
	}
	\label{cross-modal datasets}
\end{table}

\begin{table}
	\caption{Quantitative results of our SCPNet and other approaches on cross-spectral datasets. NC denotes the training is not converged. \textbf{Bold} indicates the best result among unsupervised methods.}
	\renewcommand{\tabcolsep}{4pt}{}
	\renewcommand\arraystretch{1.1}
	\centering
	\resizebox{\linewidth}{!}
	{
		\begin{tabular}{cc|cccc|cccc}
			\hline
			\specialrule{0.1em}{0em}{0em}
			\multicolumn{2}{c|}{Dataset} & \multicolumn{4}{c|}{Harvard} & \multicolumn{4}{c}{RGB/NIR} \\
			\cline{1-2}
			\multicolumn{2}{c|}{Offset} & Easy & Moderate & Hard & Mean & Easy & Moderate & Hard & Mean \\
			\hline
			\multirow{4}{*}{Handcrafted} & SIFT \cite{lowe2004distinctive} & 17.27 & 21.49 & 26.70 & 22.30 & 15.54 & 23.90 & 28.81 & 24.40 \\
			\multirow{4}{*}{} & ORB \cite{rublee2011orb} & 18.61 & 23.06 & 28.29 & 23.82 & 17.75 & 22.84 & 27.01 & 23.00 \\
			\multirow{4}{*}{} & DASC \cite{kim2015dasc} & 11.85 & 18.29 & 25.03 & 19.05 & 13.50 & 17.91 & 25.73 & 19.78 \\
			\multirow{4}{*}{} & RIFT \cite{li2019rift} & 10.41 & 15.69 & 21.98 & 16.62 & 11.22 & 13.80 & 23.34 & 16.84 \\
			\hline
			\multirow{6}{*}{Unsupervised} & UDHN \cite{nguyen2018unsupervised} & 18.03 & 22.20 & 26.55 & 22.69 & 18.54 & 23.27 & 27.16 & 23.43 \\
			\multirow{6}{*}{} & CA-UDHN \cite{zhang2020content} & 18.77 & 23.64 & 28.55 & 24.14 & 18.31 & 23.88 & 28.66 & 24.12 \\
			\multirow{6}{*}{} & biHomE \cite{koguciuk2021perceptual} & NC & NC & NC & NC & 18.61 & 23.05 & 28.18 & 23.77 \\
			\multirow{6}{*}{} & BasesHomo \cite{ye2021motion} & 19.77 & 24.20 & 28.46 & 24.57 & 19.23 & 23.44 & 28.89 & 24.41 \\
			\multirow{6}{*}{} & UMF-CMGR \cite{Wang_2022_IJCAI} & 16.61 & 21.08 & 26.25 & 21.81 & 17.04 & 22.16 & 26.53 & 22.38 \\
			\multirow{6}{*}{} & SCPNet (Ours) & \textbf{2.34} & \textbf{3.70} & \textbf{5.48} & \textbf{4.00} & \textbf{1.65} & \textbf{4.69} & \textbf{7.13} & \textbf{4.78} \\
			\hline
			\multirow{5}{*}{Supervised} & DHN \cite{detone2016deep} & 5.30 & 6.34 & 8.09 & 6.72 & 9.55 & 10.08 & 14.87 & 11.88 \\
			\multirow{5}{*}{} & MHN \cite{le2020deep} & 4.37 & 5.09 & 6.27 & 5.35 & 6.88 & 7.10 & 8.26 & 7.51 \\
			\multirow{5}{*}{} & LocalTrans \cite{shao2021localtrans} & 0.27 & 0.43 & 4.58 & 2.04 & 0.53 & 0.77 & 5.15 & 2.47 \\
			\multirow{5}{*}{} & IHN \cite{cao2022iterative} & 1.40 & 1.72 & 2.03 & 1.75 & 1.25 & 2.14 & 2.21 & 1.90 \\
			\multirow{5}{*}{} & RHWF \cite{cao2023recurrent} & 1.37 & 1.76 & 1.85 & 1.68 & 0.68 & 1.44 & 1.08 & 1.07 \\
			\hline
			\specialrule{0.1em}{0em}{0em}
		\end{tabular}
	}
	\label{cross-spectral datasets}
\end{table}

\begin{figure*}[t]
	\centering
	\includegraphics[scale=0.155]{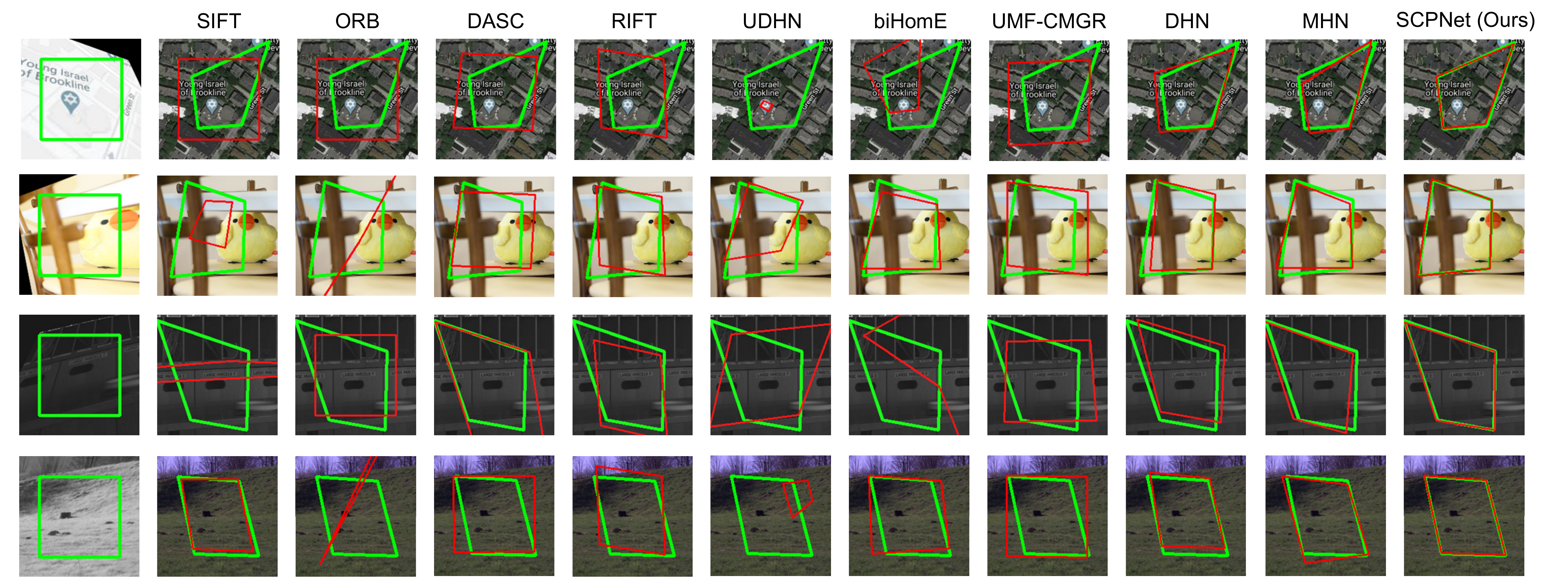}
	\caption{Qualitative homography estimation results on GoogleMap, Flash/no-flash, Harvard, and RGB/NIR datasets respectively. {\color{green}Green} polygons denote the ground-truth homography deformation from $\mathbf{I}_\mathrm{B}$ (source, the deformed image) to $\mathbf{I}_\mathrm{A}$ (target). {\color{red}Red} polygons denote the estimated homography deformation using different algorithms on the target images.}
	\label{fig:quant}
\end{figure*}

\subsection{Evaluation on Cross-modal/spectral Datasets}
We divide testing image pairs into three levels by the degree of ground-truth offsets, and define the $0\sim30\%$ as `Easy', the $30\sim60\%$ as `Moderate', and the $60\sim100\%$ as `Hard'.
Table \ref{cross-modal datasets} shows the quantitative comparison of cross-modal datasets. On GoogleMap, homography estimation faces greater challenges due to the large modality differences between image pairs. The handcrafted and other unsupervised approaches produce unsatisfactory results even under the Easy level. On the contrary, our SCPNet can produce stable and accurate homography estimation, owning 37.2\% and 14.0\% lower MACEs than the supervised approaches DHN and MHN. On Flash/no-flash, SCPNet also provides the best performance compared to all handcrafted and unsupervised approaches, and is superior to the supervised DHN and MHN. However, our SCPNet fails to suppress the supervised approaches LocalTrans, IHN and RHWF. It is due to the supervision difference and their architectures that combine iterative and multi-scale refinement.

Table \ref{cross-spectral datasets} lists the results of the cross-spectral datasets. It is observed that our SCPNet consistently surpasses other handcrafted and unsupervised approaches on Harvard and RGB/NIR datasets. On Harvard, SCPNet outperforms DHN and MHN by 40.5\%, 25.2\% respectively. We note that on Harvard dataset, the images are under intensity and gradient variation caused by the alternation of 31 spectral bands, but the training strategy of our SCPNet still works robustly. On RGB/NIR dataset, SCPNet also outperforms a part of supervised, unsupervised, and handcrafted methods as in other datasets.

Fig. \ref{fig:quant} visualizes the homography estimation results of SCPNet and other comparison approaches on GoogleMap, Flash/no-flash, Harvard, and RGB/NIR datasets. It can be seen that our SCPNet can produce accurate homography predictions on a variety of data, while the others fail due to the large modality/spectral variance and homography deformation.

\subsection{Evaluation on PDS-COCO}
\label{sec:pds-coco}
We further conduct an evaluation on PDS-COCO, with results illustrated in Table \ref{tab:pdscoco}. Following \cite{koguciuk2021perceptual}, $\delta$ represents the content distortion of brightness, contrast, saturation, and hue, the absolute value of which is bigger when there is a larger distortion. As the intensity and gradient variation of PDS-COCO dataset is inferior to the cross-modal/spectral ones, some unsupervised methods such as biHomE and UDHN produces much more accurate homography estimation than on the previous datasets. However, they are still inferior to our SCPNet. SCPNet leads biHomE by 58.2\% under the maximum content distortion and 59.5\% under the minimum one, and also outperforms the supervised methods including DHN and MHN.

\begin{table}
	\caption{Quantitative results of our SCPNet and other approaches on PDS-COCO. $\delta$ represents the content distortion of brightness, contrast, saturation, and hue. NC denotes the training is not converged. \textbf{Bold} indicates the best result among unsupervised methods.}
	\renewcommand{\tabcolsep}{2.5pt}{}
	\renewcommand\arraystretch{1.1}
	\centering
	\resizebox{\linewidth}{!}
	{
		\begin{tabular}{c|cccc|ccccc}
			\hline
			\specialrule{0.1em}{0em}{0em}
			\multirow{2}{*}{Distortion} & \multicolumn{4}{c|}{Unsupervised} & \multicolumn{5}{c}{Supervised} \\
			\multirow{2}{*}{} & UDHN & CA-UDHN & biHomE & SCPNet (Ours) & DHN & MHN & LocalTrans & IHN & RHWF \\
			\hline
			$\delta=\pm8$ & 3.24 & NC & 2.20 & \textbf{0.89} & 2.09 & 1.07 & 0.68 & 0.19 & 0.07 \\
			$\delta=\pm16$ & 5.51 & NC & 2.37 & \textbf{0.88} & 2.24 & 1.10 & 0.70 & 0.19 & 0.07 \\ 
			$\delta=\pm32$ & NC & NC & 2.61 & \textbf{1.09} & 2.50 & 1.22 & 0.79 & 0.21 & 0.09 \\
			\hline
			\specialrule{0.1em}{0em}{0em}
		\end{tabular}
	}
	\label{tab:pdscoco}
\end{table}

\subsection{Computational Burden}
The computational burden of two-branch intra-modal self-supervised learning in the training process mainly involves the extra synthetic data generation, the network forward propagation, and the computation and backward propagation of the self-supervised loss. Table.~\ref{tab:computational_burden} lists the training time and memory usage on an NVIDIA GeForce RTX 4090. Besides, we note that the inference time and memory usage will not increase.

\begin{table}
	\caption{The computational burden of training process. Self denotes intra-modal self-supervised learning.}
	\renewcommand\arraystretch{1}
	\renewcommand\tabcolsep{20pt}{}
	\centering
	{
		\begin{tabular}{c|cc}
			\hline
			\specialrule{0.1em}{0em}{0em}
			Setting & Time (Hours) & Memory usage (MBs) \\
			\hline
			w/o Self & 3.18 & 4622 \\
			w/ Self & 6.98 & 9144 \\
			\hline
			\specialrule{0.1em}{0em}{0em}
		\end{tabular}
	}
	\label{tab:computational_burden}
\end{table}

\section{Conclusions}
We have proposed a novel unsupervised cross-modal homography estimation framework, named SCPNet. The concept of intra-modal self-supervised learning is introduced for the first time, wherein two-branch self-supervised information is fully exploited by applying simulated homography within two modalities, providing strong support for unsupervised cross-modal training. Building upon this, by combining correlation and consistent feature map projection, SCPNet achieves successful unsupervised homography estimation on multiple challenging datasets. Extensive experiments demonstrate the effectiveness of SCPNet in handling large offsets and modality gaps.

\textbf{Limitations.} The homography estimation network of SCPNet is designed for better facilitating the unsupervised training framework. The strategies such as multi-scale, iteration, and replacing CNN using transformer that can further improve the homography estimation accuracy at the network design level can be further investigated in our future work.

\section*{Acknowledgments}
This work was supported in part by the National Key Research and Development Program of China under Grant No. 2023YFB3209800, in part by ``Pioneer'' and ``Leading Goose'' R \& D Program of Zhejiang under grant 2023C03136, in part by Zhejiang Provincial Natural Science Foundation of China under Grant No. LD24F020003, in part by the National Natural Science Foundation of China under Grant No. 62301484.

\bibliographystyle{splncs04}
\bibliography{manuscript}
\end{document}